\documentclass[sigconf]{acmart}

\AtBeginDocument{%
  \providecommand\BibTeX{{%
    \normalfont B\kern-0.5em{\scshape i\kern-0.25em b}\kern-0.8em\TeX}}}

\setcopyright{acmlicensed}
\copyrightyear{2024}
\acmYear{2024}
\acmDOI{10.1145/3664647.3681621}

\acmISBN{979-8-4007-0686-8/24/10}

\usepackage{subfigure}
\usepackage{colortbl}
\usepackage{color,xcolor}
\definecolor{color3}{rgb}{0.95,0.95,0.95}

\begin{document}

\title{Learning A Low-Level Vision Generalist via Visual Task Prompt}


\author{Xiangyu Chen}
\affiliation{%
  \institution{University of Macau; Shanghai AI Laboratory; Shenzhen Institute of Advanced Technology, CAS}
  \country{}
}

\author{Yihao Liu}
\affiliation{%
  \institution{Shanghai AI Laboratory; \\Shenzhen Institute of Advanced Technology, CAS}
  \country{}
}

\author{Yuandong Pu}
\affiliation{%
  \institution{Shanghai Jiao Tong University, \\Shanghai AI Laboratory}
  \country{}
}

\author{Wenlong Zhang}
\affiliation{%
 \institution{Shanghai AI Laboratory, \\The Hong Kong Polytechnic University}
 \country{}
}

\author{Jiantao Zhou}
\authornotemark[1]
\affiliation{%
  \institution{State Key Laboratory of Internet \\of Things for Smart City, \\University of Macau}
  \country{}
}

\author{Yu Qiao}
\affiliation{%
  \institution{Shanghai AI Laboratory; \\Shenzhen Institute of Advanced Technology, CAS}
  \country{}
}

\author{Chao Dong}
\authornote{Corresponding author}
\affiliation{%
  \institution{Shanghai AI Laboratory; Shenzhen Institute of Advanced Technology, CAS; Shenzhen University of Advanced Technology}
  \country{}
}

\renewcommand{\shortauthors}{Chen et al.}

\begin{abstract}
Building a unified model for general low-level vision tasks holds significant research and practical value. 
Current methods encounter several critical issues.
Multi-task restoration approaches can address multiple degradation-to-clean restoration tasks, while their applicability to tasks with different target domains (e.g., image stylization) is limited. 
Methods like PromptGIP can handle multiple input-target domains but rely on the Masked Autoencoder (MAE) paradigm. 
Consequently, they are tied to the ViT architecture, resulting in suboptimal image reconstruction quality. 
In addition, these methods are sensitive to prompt image content and often struggle with low-frequency information processing.
In this paper, we propose a Visual task Prompt-based Image Processing (VPIP) framework to overcome these challenges. 
VPIP employs visual task prompts to manage tasks with different input-target domains and allows flexible selection of backbone network suitable for general tasks.
Besides, a new prompt cross-attention is introduced to facilitate interaction between the input and prompt information.
Based on the VPIP framework, we train a low-level vision generalist model, namely GenLV, on 30 diverse tasks. 
Experimental results show that GenLV can successfully address a variety of low-level tasks, significantly outperforming existing methods both quantitatively and qualitatively. 
Codes are available at \textbf{\url{https://github.com/chxy95/GenLV}}.
\end{abstract}





\maketitle

\section{Introduction}
Low-level vision comprises a multitude of tasks that manipulate and enhance the pixel-level information of images. 
These tasks include but are not limited to image restoration, image enhancement, image feature extraction and image stylization. 
Over the years, numerous methods have been proposed to address various low-level vision tasks, many of which have achieved commendable performance for specific individual tasks~\cite{restormer,hat_journal,retinexformer}. 
However, developing task-specific models often proves to be time-consuming and labor-intensive.
Recently, there has been a significant trend in artificial intelligence towards creating general models. 
In Natural Language Processing (NLP), Large Language Models (LLMs) like the GPT series \cite{gpt3,chartgpt} have exhibited remarkable performance.
Similarly, in computer vision, models such as the Segment Anything Model (SAM) \cite{sam} and Track Anything Model (TAM) \cite{tam} have emerged, primarily focusing on high-level perceptual tasks.
However, research on general models for low-level tasks is limited.

\begin{figure*}[t]
    \centering
    \includegraphics[width=1\textwidth]{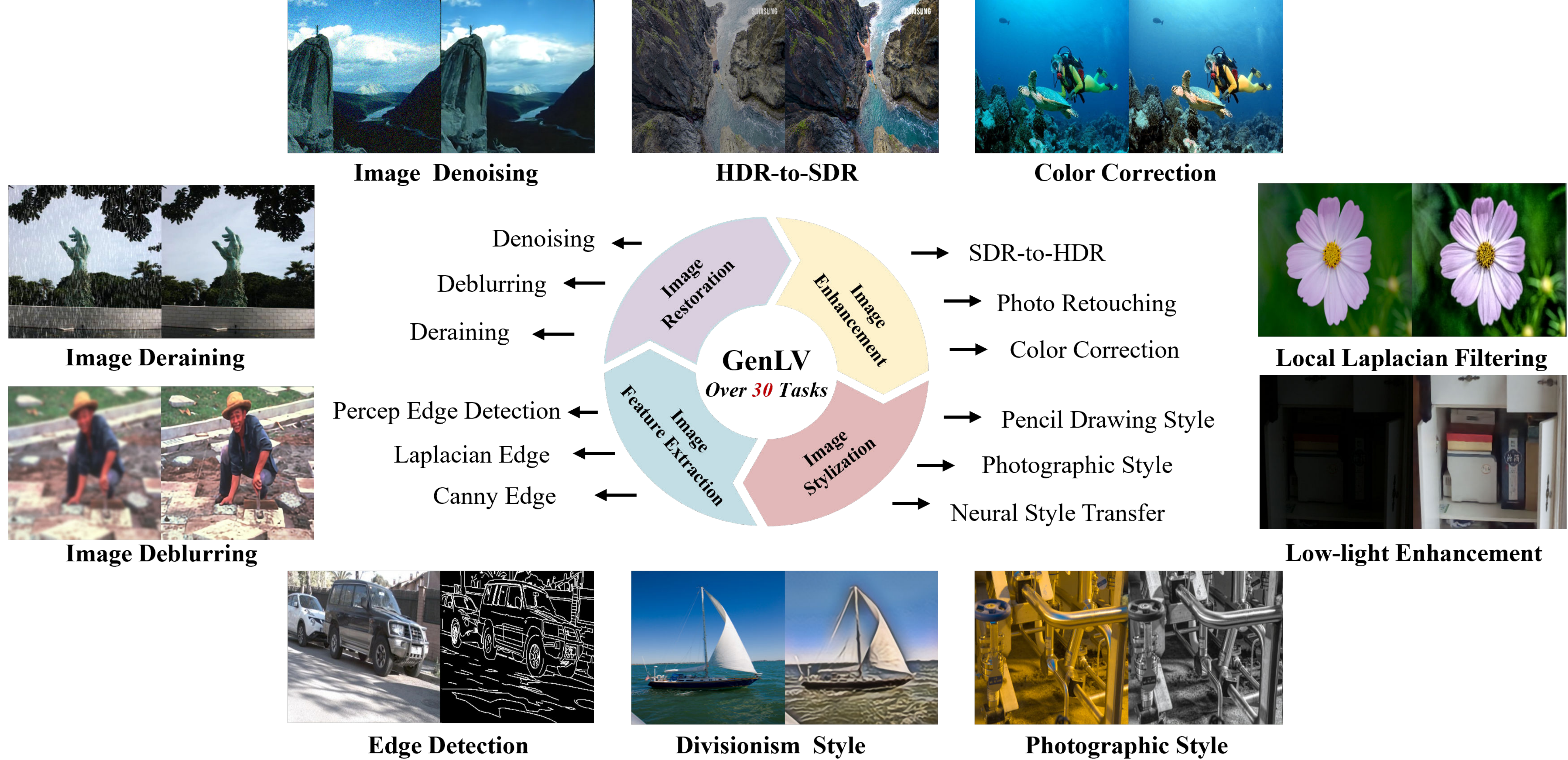}
    \vspace{-13pt}
    \caption{Our proposed low-level vision generalist model, GenLV, can handle diverse tasks with various input/target domains.}
    \label{fig:teaser}
    \vspace{-7pt}
\end{figure*} 

Designing a general model for low-level vision presents several challenges.
Firstly, the diversity of tasks involves distinct input and target domains (e.g., image restoration vs. stylization).
Unifying these within a single framework is difficult.
Existing models often target specific task categories, such as AirNet \cite{airnet} and PromptIR \cite{promptir}, which focus on restoration tasks but cannot handle feature extraction or stylization.
Secondly, for low-level vision, pixel-level reconstruction quality is crucial.
However, existing general vision models focus more on perceptual accuracy, neglecting the model’s image reconstruction capability. 
For instance, MAE-VQGAN~\cite{maevqgan} uses discrete feature representations for image reconstruction, often resulting in unacceptable structural differences between inputs and outputs~\cite{promptgip}.
Painter~\cite{painter} and PromptGIP~\cite{promptgip}, based on ViT architecture~\cite{vit}, often lack fine details and occasionally exhibit blocking artifacts. 
Furthermore, handling a wider range of low-level tasks involves processing high- and low-frequency information simultaneously, posing additional challenges.
PromptGIP~\cite{promptgip} performs well within a limited task range, but struggles when tasks increases, especially when more low-frequency information processing (e.g., color and style) is involved. 
The Painter~\cite{painter} model trained under the same low-level task settings faces similar or even more severe issues, as shown in Figure~\ref{fig:visual}. 
These challenges are attributed to the Masked Autoencoder (MAE)-based training paradigm, which makes the models sensitive to the prompt image content, especially for thelow-frequency information.
To address these issues, we propose a Visual Task Prompt-based Image Processing (VPIP) framework. It consists of three components: an end-to-end image processing network, a prompt encoder sub-network, and an information interaction mechanism for task-specific processing.
We employ the X-Restormer \cite{xrestormer} as the main network, which is designed for general image restoration tasks.
We take input-target image pairs as visual prompt to represent different tasks. 
The prompt encoder processes the visual prompt into latent representations, and a new prompt cross-attention is introduced to incorporate the latent task information into the main network.
Our VPIP framework offers several advantages: 1) It effectively solves the problem of varying input-target domains for various tasks. 2) It is not restricted to the MAE paradigm, enhancing robustness against prompt image content. 3) It allows flexibility in selecting backbone networks, improving reconstruction quality. 4) It reduces attention computation costs of global attention in previous MAE-based models by using cross-attention.

To evaluate the effectiveness of our method, we construct 30 diverse tasks to train the model. 
The trained low-level vision generalist model, namely GenLV, can successfully process the various tasks with different input and target domains, as show in Figure~\ref{fig:teaser}. 
%
%
Comprehensive experiments, detailed in Section~\ref{experiments}, demonstrate that GenLV significantly outperforms existing methods.

\section{Related Work}
\label{related_work}
\textbf{Low-Level Vision.} 
Over the past decade, low-level vision has significantly advanced due to deep learning integration. Classic tasks in this field include image restoration, enhancement, feature extraction, and stylization. 
Image restoration focuses on recovering high-quality image from degraded versions caused by factors like low-resolution~\cite{srcnn_eccv}, noise~\cite{dncnn}, blur~\cite{dpdnet}, JPEG compression~\cite{arcnn} and bad weather, such as rain~\cite{raindata} and haze~\cite{ffanet}.
Image enhancement~\cite{ImageEnhancementSurvey} involves modifying image attributes like color~\cite{hdrnet}, sharpness~\cite{llf}, exposure~\cite{hdrunet} and brightness~\cite{sidd}, to improve suitability for specific tasks or viewers.
Image feature extraction, extracts low-level features to aid downstream enhancement and understanding tasks. 
Image stylization aims to create visually appealing images with a specific style or aesthetic~\cite{perceploss}.
Despite advancements, current methods often depend on specialized datasets and customized network architectures, limiting their practical applications.

\noindent\textbf{Prompt Learning.} 
In the NLP field, the concept of prompting is initially to supply manually selected in-context information to a pretrained model for implementing the target task~\cite{gpt3}. Instead of using manual prompt, many follow-up works propose to treat the prompt as task-specific vectors to adapt model for various tasks~\cite{lester2021power,lora}. Prompt learning techniques have also been applied in computer vision, where they have proven effective in modeling task-specific instructions across various applications~\cite{vpt,zhou2022learning}. 
Notably, MAE-VQGAN~\cite{maevqgan} and Painter~\cite{painter} employ prompting to unify vision tasks, excelling in high-level tasks like semantic segmentation
However, their effectiveness in low-level vision tasks is limited~\cite{promptgip}.
In the realm of low-level vision, PromptGIP~\cite{promptgip} uses an MAE-based framework and grid-like visual prompt for 15 cross-domain tasks.
Despite this, as task complexity increases, the effectiveness of this method diminishes. 
Besides, the training paradigm adopted by PromptGIP highly relies on the ViT architecture, which greatly limits its image reconstruction quality.

\noindent\textbf{Multi-task Image Restoration.} 
Multi-task image restoration aims to train a single model to handle multiple restoration tasks simultaneously. 
Existing multi-task image restoration methods can be categorized into two groups. 
The first group of methods aim to process real-world images with unknown degradation, emphasizing the modeling of complex real-world degradation. 
The representative approaches include BSRGAN~\cite{bsrgan} and Real-ESRGAN~\cite{realesrgan}).
In contrast, the second group of methods like DASR~\cite{dasr} and AirNet~\cite{airnet} are developed to handle several specific restoration tasks with predefined degradation. 
These methods mainly focus on designing better modules for multi-task learning to maximize network capability of task-specific restoration performance. 
Some current works such as ProRes~\cite{prores} and PromptIR~\cite{promptir} are proposed to leverage a learnable prompt from the input image for better multi-task restoration. 
However, all these approaches are limited to solving the degradation-to-clean restoration problem, and lack the ability to deal with a broad range of cross-domain low-level vision tasks.
Unlike these approaches, our method aims to construct a low-level vision generalist model, which is not only capable of image restoration, but also excels at handling a wider range of cross-domain tasks, including enhancement, feature detection and stylization.

\begin{figure}
    \centering
    \includegraphics[width=1\linewidth]{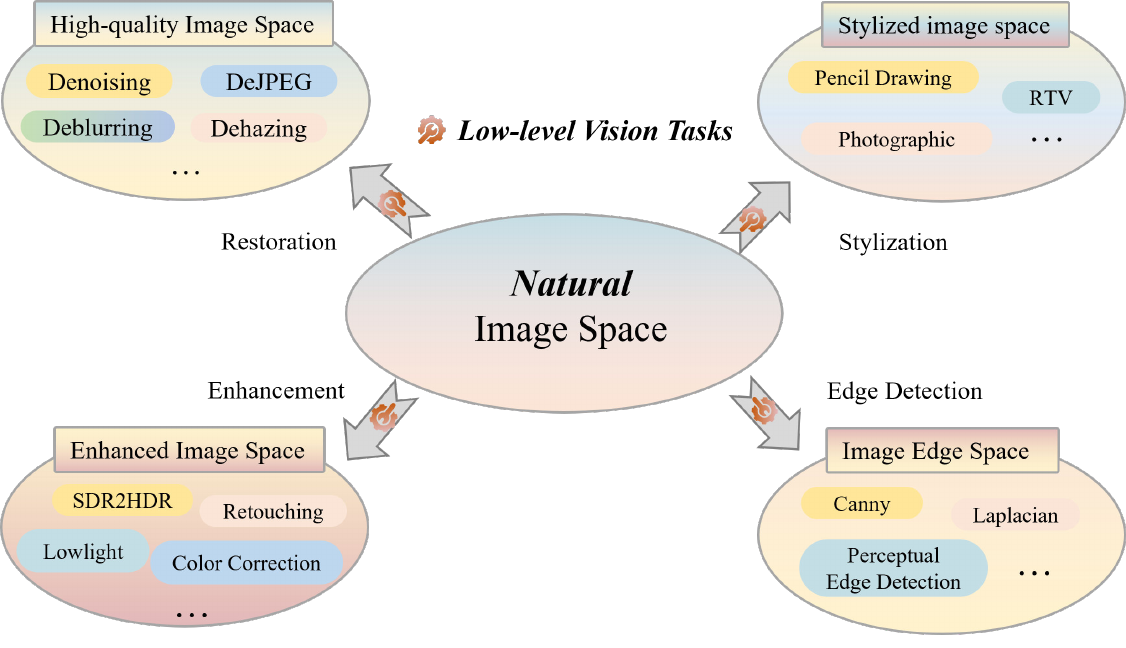}
    \caption{Diverse low-level vision tasks. Different categories of tasks differ in terms of target domains. It presents a significant challenge to build a low-level vision generalist model.} 
    \label{fig:formulation}
\end{figure} 

\section{Approach}
\label{approach}

\subsection{Representative Low-Level Vision Tasks}
Low-level vision tasks encompass a range of pixel-level manipulations, including image restoration, enhancement, feature extraction, stylization, etc. 
Each task uniquely transforms an input image space to a specific target domain. 
For example, the target domain of image restoration is high-quality (HQ) image space $\Omega_{HQ}$, while the outputs of edge detection are edges maps $\Omega_{Edge}$. 
Formally, given an arbitrary input image $I$, the low-level vision task can be defined as $\mathcal{T}_{task}: \Omega_{S} \rightarrow \Omega_{T}$, where $\Omega_{S}$ and $\Omega_{T}$ denote the source image space and the target image space, respectively. 
According to the target domain, low-level vision tasks generally fall into the following categories as: 

\begin{equation}
\begin{split}
\text{Restoration:}\quad & \mathcal{T}_{Res}: \Omega_{S} \rightarrow \Omega_{HQ}, \\
\text{Enhancement:}\quad & \mathcal{T}_{Enh}: \Omega_{S} \rightarrow \Omega_{Enh}, \\
\text{Edge Detection:}\quad & \mathcal{T}_{Edg}: \Omega_{S} \rightarrow \Omega_{Edg}, \\
\text{Stylization:}\quad & \mathcal{T}_{Sty}: \Omega_{S} \rightarrow \Omega_{Sty}.
\end{split}
\end{equation}
Each category encompasses a variety of tasks, as presented in Figure~\ref{fig:formulation}. Our goal is to address all these tasks through a unified model.

\begin{figure*}
    \centering
    \includegraphics[width=0.95\textwidth]{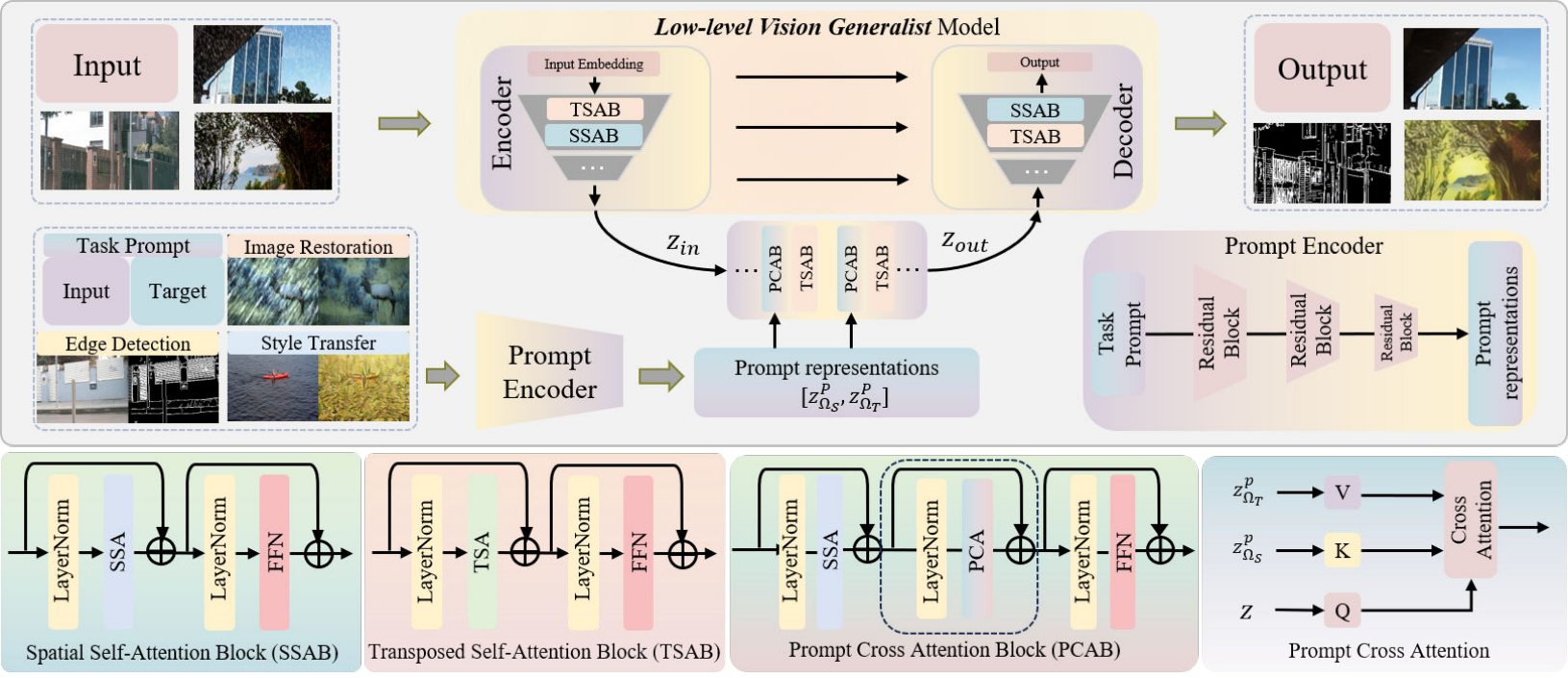}
    \caption{Overall approach of our low-level vision generalist model, GenLV.}
    \label{fig:approach}
\end{figure*} 

\subsection{Problem Formulation}
Existing low-level vision methods are typically designed for specific tasks, which inherently restricts their applicability to tasks with different target domains. 
Taking the image restoration model as an example, they accept low-quality images as input and predict the high-quality output as:
\begin{equation}
    I_{out} = \mathcal{F}_{\mathcal{T}_{Res}}(I_{in};\Theta) \in \Omega_{HQ},
\end{equation}
where $\mathcal{F}_{\mathcal{T}_{Res}}$ represents the restoration model parameterized by $\Theta$. The restoration model can accommodate various restoration tasks through incorporating multiple degradations into training, such as blur, denoising and deraining, given that the output image space of these tasks are the same, i.e., $\Omega_{HQ}$. 
However, this kind of model cannot be extended to simultaneously implement tasks like edge detection, which targets a completely different output modality. 
To train a low-level vision generalist model that can process cross-domain tasks, our approach employs a unified framework capable of handling various cross-domain tasks by utilizing additional image pair as the prompt $[P_{\Omega_{S}}, P_{\Omega_{T}}]$. It can be denoted as:
\begin{equation}
    I_{out} = \mathcal{F}_{\mathcal{T}}(I_{in}, [P_{\Omega_{S}}, P_{\Omega_{T}}]; \Theta).
    \label{single_task}
\end{equation}
This formulation allows diverse task mappings to be represented by intuitive image pairs, which marks a significant difference from conventional low-level vision models, offering a more holistic and adaptable approach to broad cross-domain low-level vision tasks. 

\subsection{Low-Level Vision Generalist Model}

In this section, we illustrate the specific design of our low-level vision generalist model, as shown in Figure~\ref{fig:approach}.
The overall approach is predicated on the Visual task Prompt-based Image Processing (VPIP) framework.
A powerful image processing network and a prompt encoder network are used to process the input image and the prompt images.
A new prompt cross-attention mechanism is introduced to achieve the information interaction among latent representations of the input image and prompt images.

\textbf{VPIP Framework} consists of an end-to-end image processing main network, a prompt encoder network and a prompt interaction mechanism. 
Given an input image $I_{in}$, it is initially processed to a high-dimensional latent feature $z_{in}$ through the encoder. 
In parallel, the paired prompt images $[P_{\Omega_{S}},P_{\Omega_{T}}]$ are fed into the prompt encoder to generate two high-dimensional representations $[z^P_{\Omega_{S}},z^P_{\Omega_{T}}]$, both of which with the same spatial size as $z_{in}$. 
Following this, the information interaction is implemented between $z_{in}$ and the pair $[z^P_{\Omega_{S}},z^P_{\Omega_{T}}]$ and results in the processed latent representation $z_{out}$. 
The final step is to reconstruct the output image from $z_{out}$ via the decoder in the main network. 
Unlike previous approaches such as Painter and PromptGIP, which rely on MAE-based framework and require binding with the ViT architecture, our VPIP framework allows flexible selection of backbone networks suitable for low-level vision tasks as the image processing main network.

\begin{figure}
    \centering
    \includegraphics[width=1\linewidth]{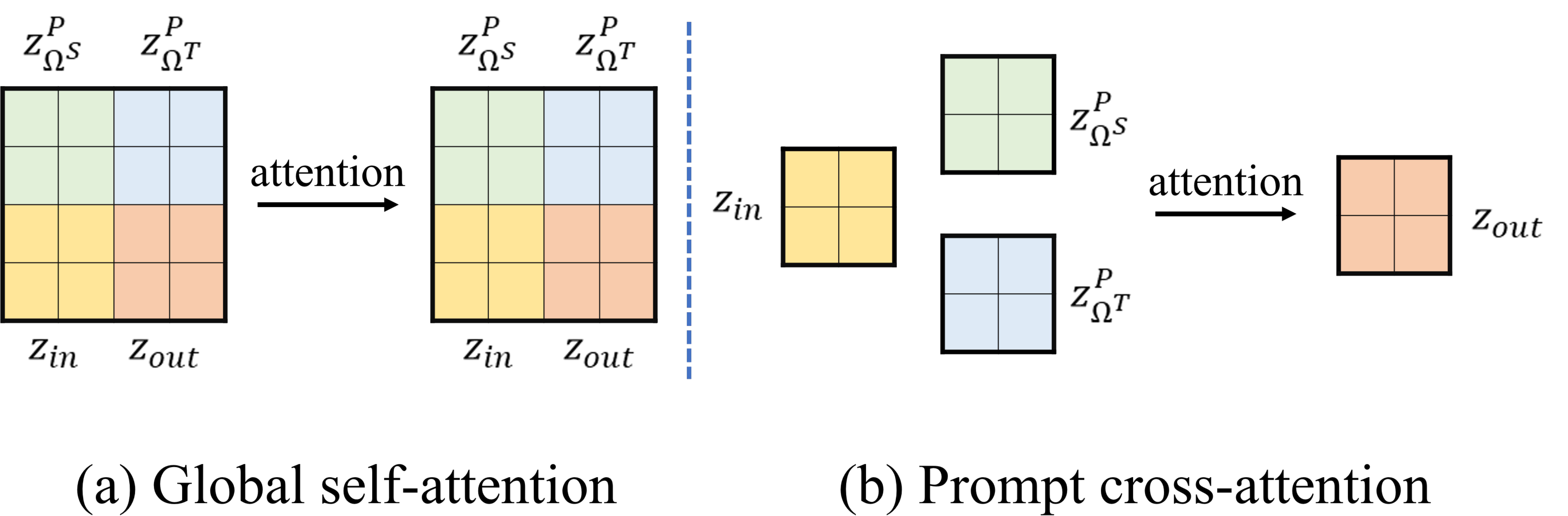}
    \caption{Comparison of two attention mechanisms.}
    \label{fig:attention}
\end{figure} 
\textbf{Image Processing Backbone} plays a crucial role in the image reconstruction quality for low-level vision tasks.
Since different low-level vision tasks often have different requirements for network capability, the most important criterion for selecting a backbone network is its task generality. 
Due to the lack of work investigating task generality across a wide range of low-level tasks, we focus more on the model performance for image restoration. 
A recent work conduct a detailed study on the backbone network for restoration tasks and propose a general backbone network, X-Restormer~\cite{xrestormer}, suitable for multiple various restoration tasks.
Therefore, we simply adopt similar architecture as our main network\footnote{We also conduct an extensive study on the model performance of different backbone network, and the detailed results are presented in the Supplementary Material.}.
Specifically, the main network employs a U-shape architecture, where downsampling and upsampling operations are performed three times, and skip connections are added from the encoder to the decoder at the same scale.
The basic modules to construct the network are Transposed Self-Attention Block (TSAB) and Spatial Self-Attention Block (SSAB), which deal with the channel-wise global information interaction and the spatial information interaction, respectively.
The structures of TSAB and SSAB are shown in Figure~\ref{fig:approach}. The implementation details of their attention mechanism can be referred to the OCA in HAT~\cite{hat} and the MDTA in Restormer~\cite{restormer}.

\textbf{Prompt Cross-attention} is designed to perform information interaction among the prompt and input representations. 
PromptGIP~\cite{promptgip} demonstrates that calculating global attention in the feature space can effectively incorporate task prompt information into image processing. 
However, this approach is tightly coupled with the ViT architecture and is relatively inefficient in terms of the attention computation. 
Inspired by the Stable Diffusion~\cite{sd} model, which utilizes cross-attention to apply text prompt to the denoising UNet, we adopt a similar mechanism to introduce visual prompt information into the image processing network.
As depicted in Figure~\ref{fig:approach}, the Prompt Cross-Attention Block (PCAB) is implemented by adding a PCA module to the standard SSAB and is integrated at the bottom of the U architecture.
To calculate the PCA, the \textit{query} (\textit{Q}), \textit{key} (\textit{K}) and \textit{value} (\textit{V}) are first generated by $1\times1$ convolutions from the input representation $z_{in}$, prompt input embedding $z^P_{\Omega_{S}}$ and prompt target embedding $z^P_{\Omega_{T}}$. 
Then, the standard attention is computed to obtain the output representation $z_{out}$.
Compared to calculating global self-attention on the grid-like features consisting of four image representations across the entire network, our prompt cross-attention calculated based on the size of one image representation in just a few blocks (i.e., PCAB) is much more efficient in terms of attention computation, as shown in Figure~\ref{fig:attention}.

\textbf{Prompt Encoder} is employed to encode the prompt images into deep representations that can be used for information interaction.
We simply utilize a series of standard residual blocks spaced by multiple downsampling operations to build the encoder network.

\section{Experiments and Analysis} 
\label{experiments} 

\subsection{Experimental Setup}
\textbf{Task Settings.} 
We train the models on \textbf{30} diverse low-level tasks, as follows: 
1) \textbf{Image Restoration:} Following PromptGIP~\cite{promptgip}, ten classic degradation types are considered including Gaussian noise, Gaussian blur, Poisson noise, salt \& pepper noise, JPEG compression, ringing artifacts, R-L algorithm~\cite{r_l}, inpainting, haze and rain. 
During the training, the on-the-fly data pairs are generated on  ImageNet~\cite{imagenet} for the first eight types. Simple mixed degradations are also considered for training. ITS dataset~\cite{RESIDE} and Rain13K~\cite{rain13k} are utilized for dehazing and deraining. A simple additive rain model is also employed for synthetic data. For testing, a mixed dataset, Common528~\cite{promptgip}, composed of several low-level vision benchmark datasets is employed. 
2) \textbf{Image Enhancement:} This category includes eight tasks: low-light enhancement (LLE), photo retouching, local Laplacian filtering~\cite{llf} (LLF), multi-scale tone manipulation (MTM)~\cite{multitone}, underwater image contrast enhancement (ICE) based on histogram equalization, underwater image color correction (ICC) based on the DIVE+ software, image SDR-to-HDR and HDR-to-SDR~\cite{hdrtvnet}. 
The LOL dataset~\cite{lol} is used for LLE, and expert-C retouched images of the Adobe-MIT Fivek dataset~\cite{fivek} are used for retouching, LLF, and MTM. The UIEB dataset~\cite{uiebd} is utilized for the two underwater image enhancement tasks. 
3) \textbf{Image Edge Detection:} This category includes three edge detection tasks: Canny operator, Laplacian operator and a perceptual edge detection (PED)~\cite{ped}.
4) \textbf{Image Stylization:} Nine style are chosen, including pencil drawing~\cite{pencildraw}, photographic style~\cite{llf}, relative total variation (RTV)~\cite{rtv}, Vermeer style, JOJO style, Raphael style, Fauvism style, Divisionism style and Cloisonnism style. Expert-C retouched images of Adobe-MIT Fivek dataset~\cite{fivek} are also used to generate the image pairs. Data of the first three styles are implemented via available toolkit, and the last six neural styles are generated by a state-of-the-art style transfer method AdaAttN~\cite{adaattn}. 

\vspace{1pt}
\noindent\textbf{Implement details.} For the backbone network, we adopt the similar setting as the original X-Restormer~\cite{xrestormer}. From level-1 to level-4, the numbers of consecutive blocks (each block contains a TSAB and an SSAB) are [2, 4, 4, 4], attention heads in TSA and SSA are both [1, 2, 4, 8], and channel numbers are [48, 96, 192, 384]. For the prompt encoder network, we employ four residual blocks for each downsampling level. During the training, the input size is set to $256\times256$ for the input image and prompt images. $L_1$ loss is utilized as the loss function. AdamW optimizer with $\beta_1=0.9$ and $\beta_2=0.99$ is adopted with an initial learning rate of $1e^{-4}$. The batch size is set to 64 and total training epochs are 30. 

\subsection{Quantitative Results}

\begin{table*}[!t]
\centering
\caption{Quantitative results on image restoration tasks. $\#$: public released model. $\star$: trained with only restoration tasks. $\dagger$: trained with all 30 low-level vision tasks. GN: Gaussian noise. PN: Poisson noise. ViT-VPIP: ViT backbone adopted in the VPIP framework. Our GenLV can also be represented as X-Restormer-VPIP. PSNR$\uparrow$ (dB) is calculated as the quantitative metric.}
\label{tab:restoration}
\resizebox{\linewidth}{!}{
\begin{tabular}{c|ccccccccccc}
\toprule
\rowcolor{color3} & GN & PN & S\&P Noise & GB & JPEG & Ringing & R-L & Inpainting & SimpleRain & ComplexRain & Haze     \\ 
\midrule
Real-ESRGAN\textsuperscript{$\#$}          & 25.38 & 26.57 & 21.50 & 21.49 & 25.21 & 24.64 & 21.71 & 14.06 & 16.10 & 21.01 & 11.86 \\
PromptIR\textsuperscript{$\star$}          & 28.86 & 31.48 & 36.45 & 24.56 & 26.77 & \textbf{27.85} & 31.31 & \textbf{28.11} & 30.76 & 24.08 & 16.85 \\ 
PromptGIP\textsuperscript{$\star$}          & 26.48 & 27.76 & 28.08 & 22.88 & 25.86 & 25.69 & 27.05 & 25.28 & 25.79 & 24.33 & 24.55 \\ 
ViT\textsuperscript{$\star$}               & 24.67 & 25.39 & 23.71 & 22.17 & 24.76 & 23.89 & 24.09 & 23.11 & 23.21 & 23.04 & 24.91 \\
ViT-VPIP\textsuperscript{$\star$}           & 26.14 & 27.20 & 25.43 & 24.13 & 26.19 & 25.98 & 26.98 & 25.03 & 25.51 & \textbf{24.79} & 24.06 \\
X-Restormer\textsuperscript{$\star$}       & 28.70 & 31.36 & 35.33 & 24.13 & 26.68 & 26.88 & 30.01 & 27.68 & 29.65 & 24.39 & 16.73 \\
GenLV\textsuperscript{$\star$} (ours)            & \textbf{28.99} & \textbf{31.69} & \textbf{36.63} & \textbf{24.58} & \textbf{26.91} & 27.74 & \textbf{31.50} & \textbf{28.11} & \textbf{31.10} & 24.71 & \textbf{28.91} \\
\midrule    
Painter\textsuperscript{$\dagger$}      & 24.28 & 24.41 & 24.93 & 21.55 & 22.30 & 23.58 & 24.36 & 22.52 & 22.42 & 23.14 & 20.20 \\
PromptGIP\textsuperscript{$\dagger$}    & 23.63 & 23.98 & 25.05 & 20.84 & 22.21 & 23.86 & 24.94 & 22.11 & 23.16 & 21.79 & 21.90 \\
ViT-VPIP\textsuperscript{$\dagger$}    & 25.30 & 26.15 & 24.41 & 22.74 & 25.35 & 24.62 & 25.24 & 23.73 & 24.00 & 23.70 & 24.04 \\
GenLV\textsuperscript{$\dagger$} (ours) & \textbf{28.49} & \textbf{31.05} & \textbf{34.20} & \textbf{23.39} & \textbf{26.21} & \textbf{25.78} & \textbf{28.21} & \textbf{27.17} & \textbf{28.18} & \textbf{25.11} & \textbf{29.70} 
\\ 
\bottomrule
\end{tabular}
}
\end{table*}

\begin{table*}[!t]
\centering
\caption{Quantitative results on image enhancement and stylization tasks. PSNR$\uparrow$ (dB) is calculated as the quantitative metric.}
\label{tab:enhance_trans}
\resizebox{\linewidth}{!}{
\begin{tabular}{c|ccccccccccc}
\toprule
\rowcolor{color3} & LowLight & LLF& Retouching & ICC & ICE & MTM & SDR2HDR & HDR2SDR & PencilDraw & Photographic & RTV     \\ 
\midrule
Painter\textsuperscript{$\dagger$}      & 20.19 & 23.98 & 18.29 & 21.62 & 15.89 & 21.51 & 25.63 & 20.56 & 16.79 & 22.68 & 26.69 \\
PromptGIP\textsuperscript{$\dagger$}    & 18.60 & 25.40 & 20.44 & 24.29 & 16.16 & 20.84 & 26.40 & 18.87 & 17.74 & 21.68 & 30.29 \\
ViT-VPIP\textsuperscript{$\dagger$}      & 22.16 & 23.78 & 22.01 & 27.70 & 16.86 & 26.10 & 27.89 & 23.91 & 19.56 & 22.30 & 31.89 \\
GenLV\textsuperscript{$\dagger$} (ours) & \textbf{23.55} & \textbf{27.61} & \textbf{23.84} & \textbf{35.44} & \textbf{17.36} & \textbf{31.59} & \textbf{34.45} & \textbf{35.92} & \textbf{20.00} & \textbf{23.86} & \textbf{33.03} 
\\ 
\bottomrule
\end{tabular}
}
\end{table*}

\begin{table}[!t]
\centering
\caption{Quantitative results on edge detection tasks. Mean absolute error$\downarrow$ is calculated as the quantitative metric.}
\label{tab:edge_dec}
\setlength{\tabcolsep}{5.8mm}{
\resizebox{1\linewidth}{!}{
\begin{tabular}{c|ccc}
\toprule
\rowcolor{color3} & Canny & Laplacian & PED \\ 
\midrule
Painter\textsuperscript{$\dagger$}      & 31.36 & 7.06 & 9.55 \\
PromptGIP\textsuperscript{$\dagger$}    & 19.48 & 4.06 & 9.36 \\
ViT-VPIP\textsuperscript{$\dagger$}      & 27.68 & 5.49 & 8.44 \\
GenLV\textsuperscript{$\dagger$} (ours) &  \textbf{8.07} & \textbf{1.27} & \textbf{7.23} \\ 
\bottomrule
\end{tabular}
}
}
\end{table}

The quantitative results for various low-level vision tasks are presented in Table~\ref{tab:restoration}, Table~\ref{tab:enhance_trans} and Table~\ref{tab:edge_dec}. Given that not all existing methods are capable to handle tasks across different target domains, our primary comparative experiments are centered on restoration tasks in Table~\ref{tab:restoration}. 
We consider three distinct experimental settings. The first setting utilizes a pretrained model, i.e., Real-ESRGAN~\cite{realesrgan}, which is capable of handling a variety of restoration tasks. The second setting is based on the training configuration outlined in our paper, but it solely focuses on image restoration tasks. The third setting involves training on the all 30 low-level vision tasks.

\noindent\textbf{Ablation Study on Visual Prompt.} 
Since the models without using task prompt cannot process tasks with different target domains, we conduct the ablation study of the visual task prompt on restoration tasks. 
In Table~\ref{tab:restoration}, ViT\textsuperscript{$\star$} and X-Restormer\textsuperscript{$\star$} are two end-to-end models only trained on image restoration tasks, while ViT-VPIP\textsuperscript{$\star$} and GenLV\textsuperscript{$\star$} (the GenLV model can also be represented as X-Restormer-VPIP) are models based on our VPIP framework, utilizing ViT and X-Restormer as their backbone respectively. 
Upon the incorporation of prompt learning, both ViT-VPIP\textsuperscript{$\star$} and GenLV\textsuperscript{$\star$} exhibit substantial performance gains over ViT\textsuperscript{$\star$} and X-Restormer\textsuperscript{$\star$} in most restoration tasks. 
This demonstrates the effectiveness of the visual prompt in facilitating the backbone network to better handle various tasks.
It is noteworthy that X-Restormer\textsuperscript{$\star$}, without using visual prompt, struggles with the dehazing task, achieving only 16.73dB. 
A similar phenomenon also occurs for the multi-task restoration method PromptIR~\cite{promptir}. 
In contrast, GenLV\textsuperscript{$\star$} tackles it considerably better, reaching 25.63dB. All these results show the effectiveness of our proposed VPIP framework.

\noindent\textbf{Influence of Backbone Network.} 
In Table~\ref{tab:restoration}, when trained on the same setting, the performance of models using X-Restormer as the backbone network (i.e., X-Restormer\textsuperscript{$\star$}, GenLV\textsuperscript{$\star$} and GenLV\textsuperscript{$\dagger$}) significantly surpasses that of models using ViT (i.e., ViT\textsuperscript{$\star$}, ViT-VPIP\textsuperscript{$\star$} and ViT-VPIP\textsuperscript{$\star$}).
This observation suggests that an appropriate backbone network is important for low-level vision tasks generalist models, and ViT architecture may limit the model performance. 

\noindent\textbf{Comparison with other methods.} 
In Table~\ref{tab:restoration}, GenLV\textsuperscript{$\star$} outperforms the state-of-the-art blind SR method Real-ESRGAN~\cite{realesrgan} and multi-task restoration method PromptIR~\cite{promptir}, when only considering image restoration tasks. 
Note that we retrain the PromptIR model on the same setting for fair comparison (the original PromptIR is trained only on 4 tasks). 
By employing the ViT network, PromptGIP\textsuperscript{$\star$} trained on restoration tasks performs better than ViT-VPIP\textsuperscript{$\star$}, due to more attention computation. However, as more tasks are involved, ViT-VPIP\textsuperscript{$\dagger$} outperforms PromptGIP\textsuperscript{$\dagger$} and Painter\textsuperscript{$\dagger$} instead, showing the superiority of our framework for solving more diverse tasks. 
In Table~\ref{tab:enhance_trans} and Table~\ref{tab:edge_dec}, we further show the quantitative results on broader low-level vision tasks. 
Only methods capable of solving tasks across different target domains are considered in the comparison. 
The models employed VPIP framework outperform Painter and PromptGIP on a variety of low-level vision tasks. 

\vspace{-2.5pt}
\subsection{Visual Results}

\begin{figure*}
    \centering
    \includegraphics[width=0.95\linewidth]{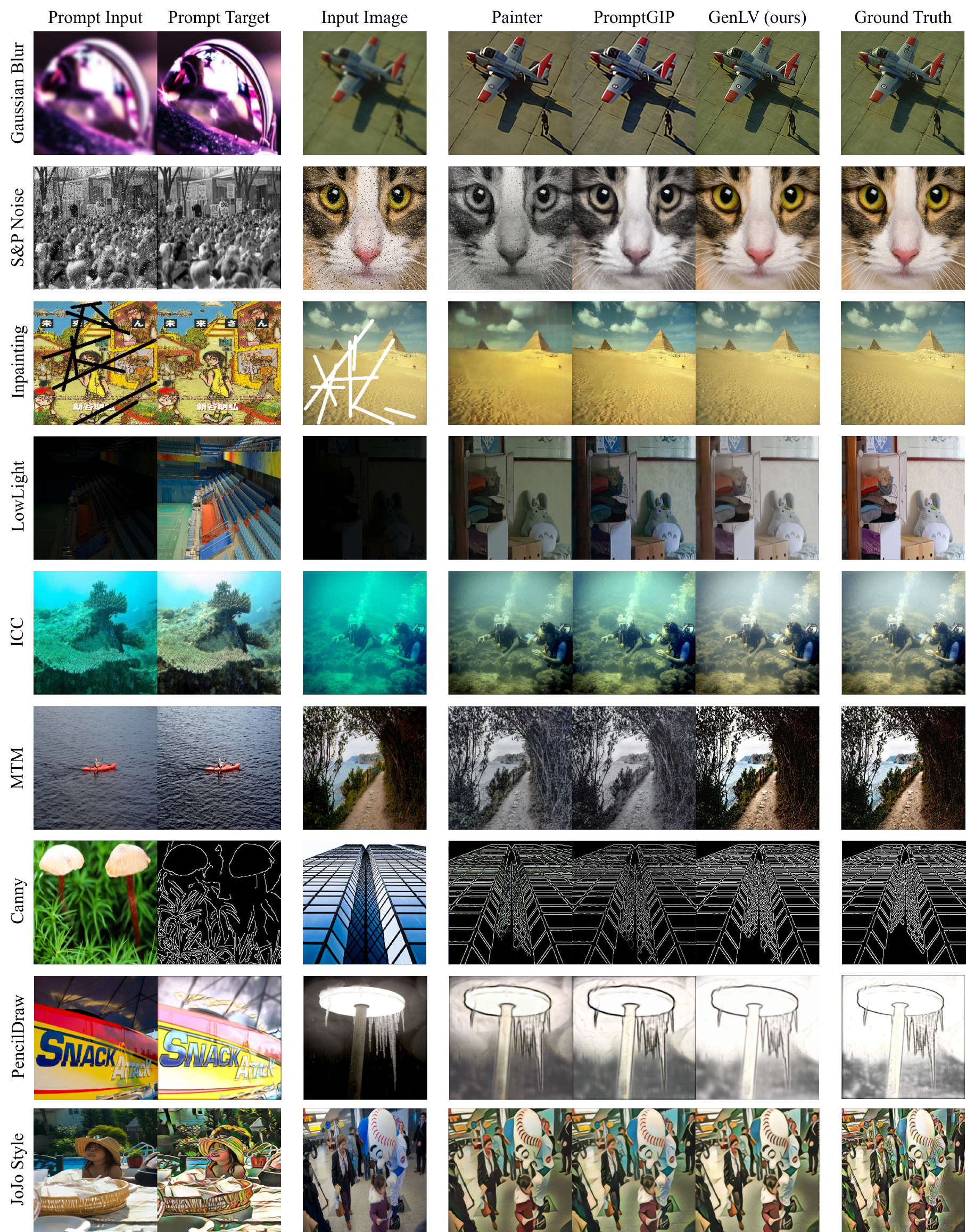}
    \caption{Visual results of different models on various low-level vision tasks.}
    \label{fig:visual}
\end{figure*}

In Figure~\ref{fig:visual}, we present the visual comparison of our GenLV with Painter and PromptGIP across various low-level vision tasks. 
From a holistic perspective, GenLV produces results that are more consistent with the ground truth, especially in aspects such as color and brightness.
In contrast, the results produced by Painter and PromptGIP are easily affected by errors in low-frequency information, manifesting as color anomalies or even incorrect task execution.
Rather that our method where prompt information can accurately serve as task instruction, Painter and PromptGIP appear to be significantly affected by the content of the prompt image. 
In terms of image reconstruction quality, the images generated by GenLV have clear textures and details. 
Conversely, Painter and PromptGIP may suffer from blurring or blocking artifacts, particularly for image restoration tasks.
Overall, the above results show the superiority of GenLV in visual quality for dealing with various low-level tasks.

\subsection{Exploration of Task Prompt} The above results have demonstrated the advantages of our prompt mechanism compared to existing methods from quantitative and qualitative perspectives. In this section, we conduct more experiments to further illustrate the effectiveness and explore the limitation of the task prompt in our method. 

\vspace{2pt}
\noindent\textbf{Influence of Different Prompts.} To explore the influence on the quantitative performance for different prompt images, we randomly select 20 prompt image pairs for each task and calculate the performance on the corresponding test sets. Then, we compute the standard deviation of the 20 performance results for each task, as shown in Table~\ref{tab:std}. 
We can see that except for PencilDraw, the standard deviations are around or lower than 0.1dB. This shows that our method is stable in performance for different prompts. 

\begin{table}[!t]
\centering
\caption{Standard deviation of the performance computed based on 20 different prompt images. PSNR (dB) is calculated as the quantitative metrics.}
\vspace{-1.6pt}
\label{tab:std}
\setlength{\tabcolsep}{0.8mm}{
\resizebox{1\linewidth}{!}{
\begin{tabular}{c|ccccccc}
\toprule
\rowcolor{color3} & GN & GB & LowLight & ICC & PencilDraw & RTV \\ 
\midrule
Painter\textsuperscript{$\dagger$}   & 2.3930 & 1.8845 & 1.8865 & 1.9573 & 1.1820 & 2.6163\\
PromptGIP\textsuperscript{$\dagger$} & 3.1035 & 2.2893 & 0.6766 & 0.6311 & 1.4200 & 1.3130 \\
GenLV\textsuperscript{$\dagger$}     & \textbf{0.1033} & \textbf{0.0208} & \textbf{0.0399} & \textbf{0.0512} & \textbf{0.5518} & \textbf{0.0195} \\
\bottomrule
\end{tabular}
}
}
\end{table}

\begin{figure}[!t]
\centering
\vspace{-4.5pt}
\subfigure[Results for mixed degraded images.]{
\label{fig:mix}
\begin{minipage}[t]{1\linewidth}
\centering
\includegraphics[width=1\linewidth]{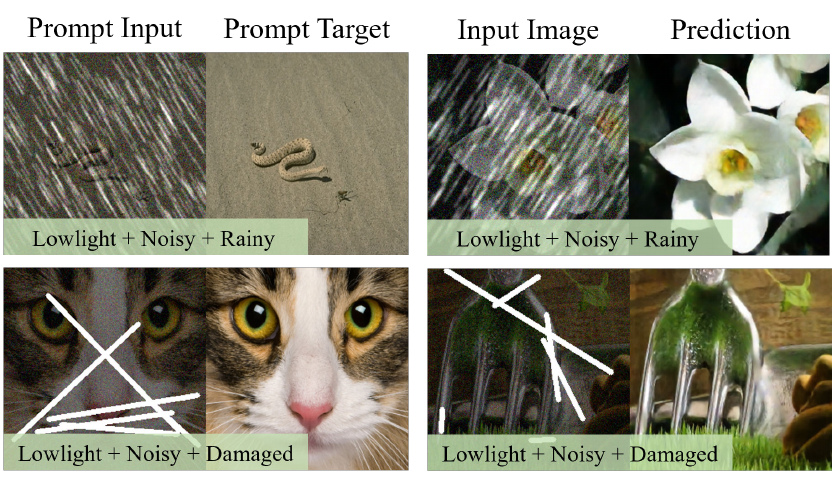}
\end{minipage}%
}

\vspace{-3pt}
\subfigure[Results for images based on cross-domain prompts.]{
\label{fig:domain}
\begin{minipage}[t]{1\linewidth}
\centering
\includegraphics[width=1\linewidth]{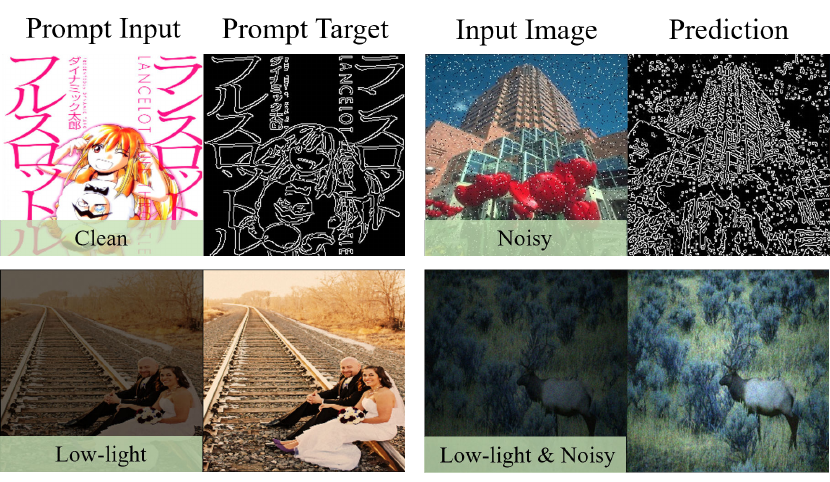}
\end{minipage}%
}

\vspace{-3pt}
\subfigure[Results for mixed degraded images on single-task prompts.]{
\label{fig:guidance}
\begin{minipage}[t]{1\linewidth}
\centering
\includegraphics[width=1\linewidth]{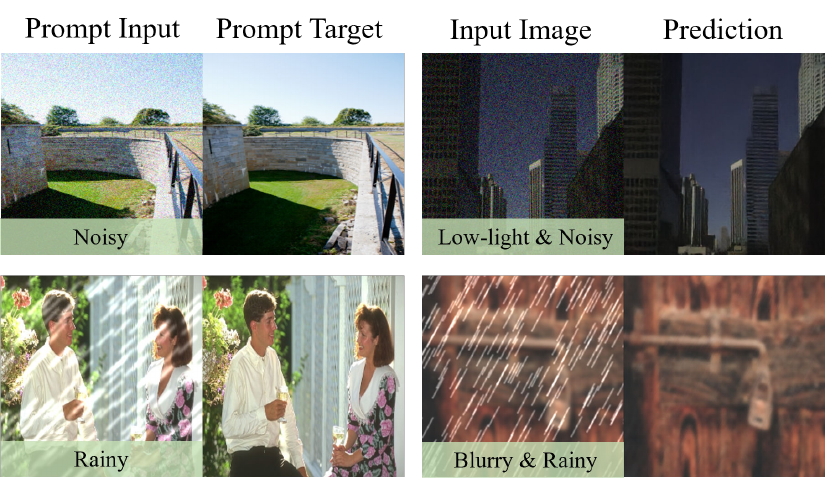}
\end{minipage}%
}
\vspace{-3pt}
\caption{Results for task prompts on complex situations.}
\vspace{-5pt}
\label{fig:complex}
\end{figure}

\vspace{2pt}
\noindent\textbf{Task Prompt on Complex Situations.} We conduct further experiments to investigate the effectiveness of task prompt on complex situations. 
In Figure~\ref{fig:mix}, we exhibit the outputs for images subjected to mixed degradation. The results show that the task prompt successfully guide the mapping under this situation, and our method has the capability to deal with tasks with mixed degradation. 
In Figure~\ref{fig:domain}, we present the results for cross-domain prompt. Utilizing Canny edge detection and LLE prompts, we instruct the model to process the noisy images. We can see that our model accurately execute the target task according to the visual prompts other than perform denoising. 
In Figure~\ref{fig:guidance}, we show the results on processing mixed degraded images using single-task prompts. The first row present the application of a denoising prompt to a low-light, noisy image. In the second row, we show that a deraining prompt is applied to a blurry image rain streaks. It can be see that the task-specific prompts effectively guide the model to perform the target task. 
All these results demonstrate the effectiveness of the visual task prompt in our method across a variety of complex situations.

\begin{figure}[!t]
    \centering
    \includegraphics[width=0.98\linewidth]{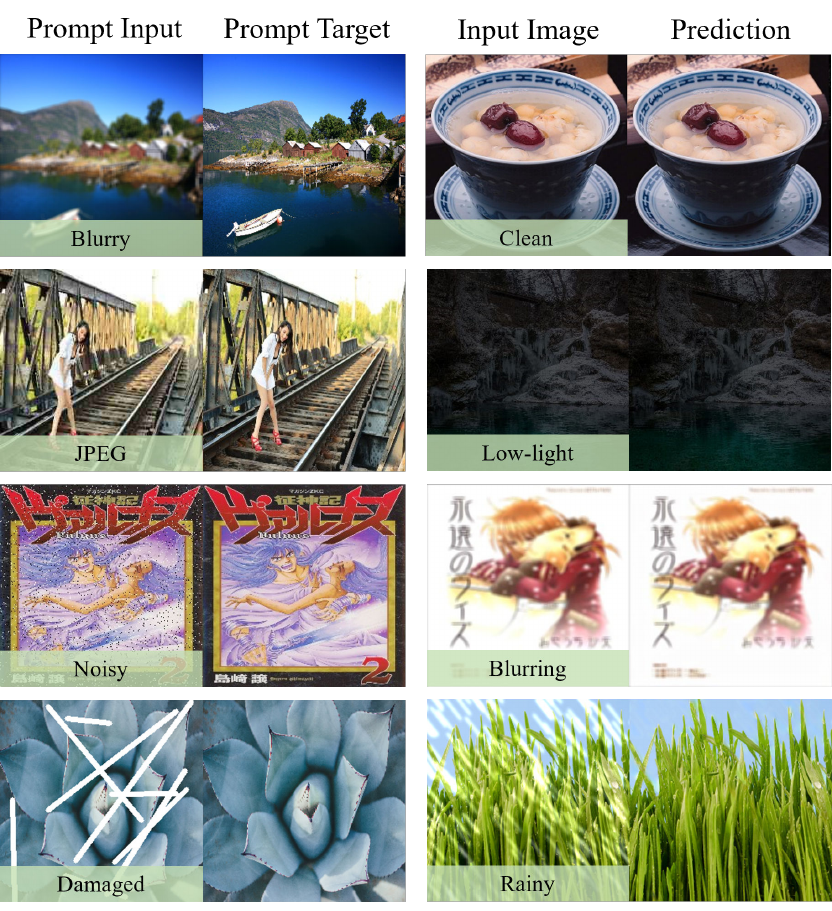}
    \vspace{-1pt}
    \caption{Results of the mismatch test.}
    \vspace{-4pt}
    \label{fig:mismatch}
\end{figure}

\vspace{2pt}
\noindent\textbf{Mismatch Test.} We conduct mismatch test to illustrate the impact of the prompt on the model under special scenarios, as shown in Figure~\ref{fig:mismatch}. 
The first row demonstrates providing the deblurring prompt to a clean image. 
In the second and third rows, we provide deJPEG and denoising prompts for low-light and blurry images, respectively. 
Ideally, we hope that the model do not execute the wrong prompt (this is reasonable from the perspective of the prompt cross-attention mechanism). 
It is observed that the model ideally preserves the original input images instead of performing degradation removal in these three instances. 
However, the mismatch test does not consistently yield ideal outcomes. 
In the fourth row, the model conducts deraining when provided with an inpainting prompt.
From this perspective, this indicates that the model still inevitably overfit some data or mappings during training.


\section{limitations and Prospects} 
Our GenLV model demonstrates commendable performance in solving a broad range of low-level vision tasks, leveraging the visual prompt-based image processing framework and a powerful backbone network. 
Nonetheless, there are certain limitations and potential areas for further exploration that warrant attention. 
The working mechanism of this method is still to divide the task space via visual prompt, to achieve multi-task low-level vision.
Despite the considerable improvement in performance compared to existing methods, we have to claim that the model currently still lacks the ability to generate satisfactory results for out-of-distribution unseen tasks.
Recent study about large language models (LLMs) underscore that the effectiveness of LLMs largely depends on the quality, diversity, and quantity of the training data~\cite{wei2023larger}. 
However, the task variety, model size (\textasciitilde30M), data scale (\textasciitilde140W) are not sufficient for GenLV. We hope that future studies involving larger models, broader tasks and data will yield more surprising results.

\section{Conclusion}
In this paper, we introduce a low-level vision generalist model, GenLV, which is capable of addressing various low-level vision tasks. Our approach involves the design of an image processing framework based on visual task prompt, VPIP, which enables the model to accommodate multiple tasks with different target domains. In addition, this framework allows the flexibility to incorporate a powerful backbone network that is suitable for low-level vision tasks, resulting in superior image reconstruction quality. Experimental results demonstrate that our GenLV can effectively manage 30 diverse low-level vision tasks and significantly outperform existing methods quantitatively and qualitatively.

\clearpage

\begin{acks}
This work was partially supported by National Natural Science Foundation of China (Grant No.62276251, 62272450), and the Joint Lab of CAS-HK. This work was also supported in part by Macau Science and Technology Development Fund under SKLIOTSC-2021-2023 and 0022/2022/A. 
\end{acks}

\bibliographystyle{ACM-Reference-Format}
\bibliography{acmmm}

\appendix

\vspace{20pt}
\section{Exploration on Different Image Restoration Backbone Networks}
The quantitative comparison of various backbone networks for different image restoration tasks is presented in Table~\ref{tab:backbone}. 
All the models are trained on the same multi-task restoration setting.
We explore the different backbone networks based on image restoration because it has a clear quantitative evaluation
scheme (i.e., PSNR/SSIM) and numerous low-level vision networks are designed based on it.
As one can see that the overall performance of X-Restormer is the best, so it is selected as the backbone network in our method.
It is noteworthy that the dehazing results show unusual performance, as both Restormer and X-Restormer perform much worse over the other comparison networks on this task.
After inspecting the results, we find that these two models do not process some haze images. 
This suggests that these two networks may have fatal optimization difficulty in handling multi-task image restoration when dehazing is considered. 
Nevertheless, we can see that the introduction of task prompts effectively mitigates this problem, as depicted in Table 1 of the main paper. 

\vspace{5pt}
\section{Exploration on Different Prompt Interaction mechanisms}
In this section, we explore the impact of different prompt interaction mechanisms in the proposed VPIP framework.
A model without using prompt and two models using common modulation strategies (i.e., global feature modulation (GFM)~\cite{csrnet} and spatial feature transform (SFT)~\cite{sftgan}) for low-level vision tasks are compared.
All models are trained on the same settings involving 30 tasks. 
In Table~\ref{tab:modulation}, we present the quantitative results of these models on restoration tasks.
We can see that the model without using prompt performs much worse than other models. 
This is reasonable because the model cannot handle tasks with different target domains (e.g., edge detection), which greatly affects the optimization. 
Models with GFM and SFT can achieve much better performance than the model without prompt interaction, but their performance is still lower than our model. 
This suggests that the feature modulation schemes can also achieve task guidance to a certain extent, but their ability to learn the 
task representation is not as effective as prompt cross-attention. 

\vspace{5pt}
\section{Comprehensive comparison between PromptGIP and our GenLV}
We conduct comprehensive experiments and demonstrate the quantitative comparison of PromptGIP and GenLV under three training settings.
Trained only for restoration tasks, we can see that our GenLV\textsuperscript{$\star$} can already outperforms PromptGIP\textsuperscript{$\star$}. 
This is mainly due to the powerful backbone that our VPIP framework can use.
When the number of tasks increases to 15 (i.e., the PromptGIP setting), the performance of both PromptGIP\textsuperscript{$\#$} and GenLV\textsuperscript{$\#$} decreases slightly, while no more than 0.5dB.
As the complexity of tasks continues to increase (i.e., the GenLV setting), we can find that the performance of both PromptGIP\textsuperscript{$\dagger$} and GenLV\textsuperscript{$\dagger$} drops significantly.
However, the performance degradation of GenLV\textsuperscript{$\dagger$} on most tasks is within 1dB, while PromptGIP\textsuperscript{$\dagger$}'s performance degradation is around 2 to 4dB. 
This intuitively indicates that PromptGIP is more easily affected by the increase in the number and complexity of tasks. 
From the main paper, we illustrate that this is because PromptGIP is sensitive to the prompt content.
When more tasks involving low-frequency processing are considered, its performance would be greatly affected. 
Many cases in the visual comparison (in the main paper Figure 5, Supp. Figure 1, 2 and 3 ) can more directly reflect this point.

\vspace{5pt}
\section{Computation Cost Breakdown}
In Table~\ref{tab:computation}, we present the computation cost of different components of our GenLV model.
The computational cost of the main network is similar to the original X-Restormer.
The computational cost of prompt encoder comes from the residual blocks.
The computational cost of the extra fusion part is from the prompt cross-attention modules in PCABs. 
Our prompt interaction scheme can bring considerable performance improvement at limited additional cost.

\begin{table}[h]
\centering
\vspace{-3pt}
\caption{Computation cost of different parts of our model.}
\label{tab:computation}
\vspace{-3pt}
\setlength{\tabcolsep}{7mm}{
\resizebox{1\linewidth}{!}{
\begin{tabular}{ccc}
\toprule
\rowcolor{color3} Component & Params & MACs \\ 
\midrule
Main Network & 27.6M & 166.9G  \\
Prompt Encoder \& Fusion & 5.1M & 35.7G  \\
\bottomrule
\end{tabular}
}
}
\end{table}

\section{More Visual Results}
\begin{figure*}[h]
    \centering
    \vspace{0.9cm}
    \includegraphics[width=1\textwidth]{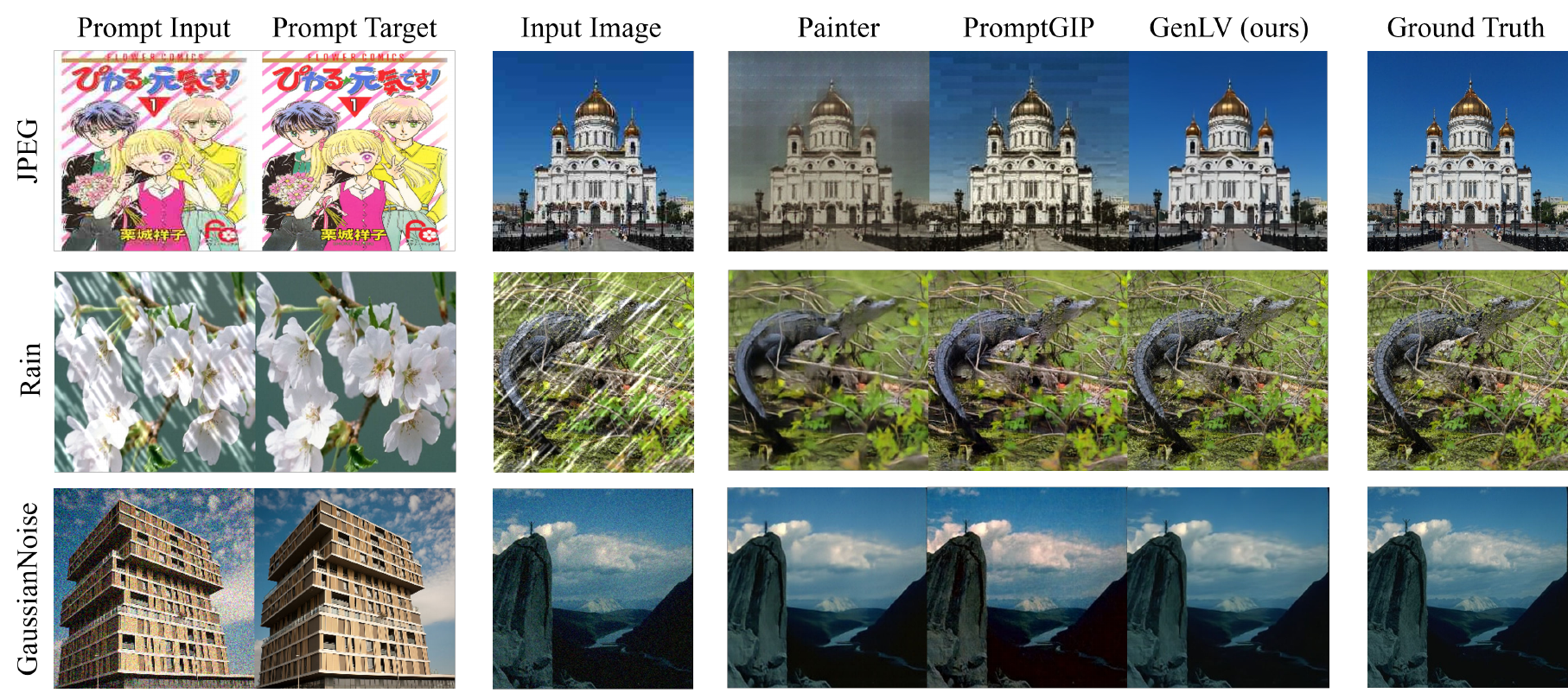}
    \caption{More visual results of different models on various low-level vision tasks.}
    \label{fig:visual2}
\end{figure*}

\begin{figure*}
    \centering
    \includegraphics[width=0.95\textwidth]{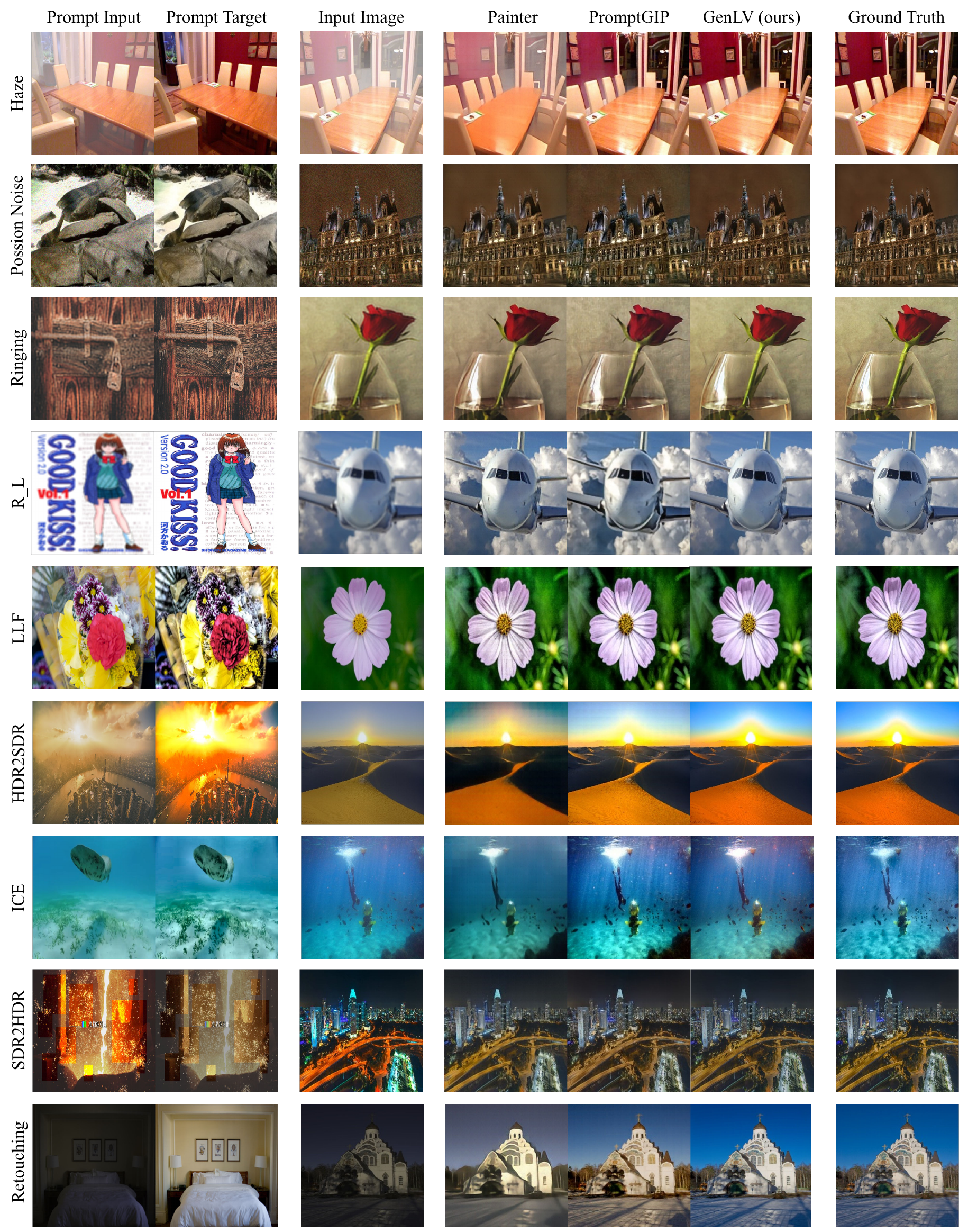}
    \caption{More visual results of different models on various low-level vision tasks.}
    \label{fig:visual3}
\end{figure*}

\begin{figure*}
    \centering
    \includegraphics[width=0.95\textwidth]{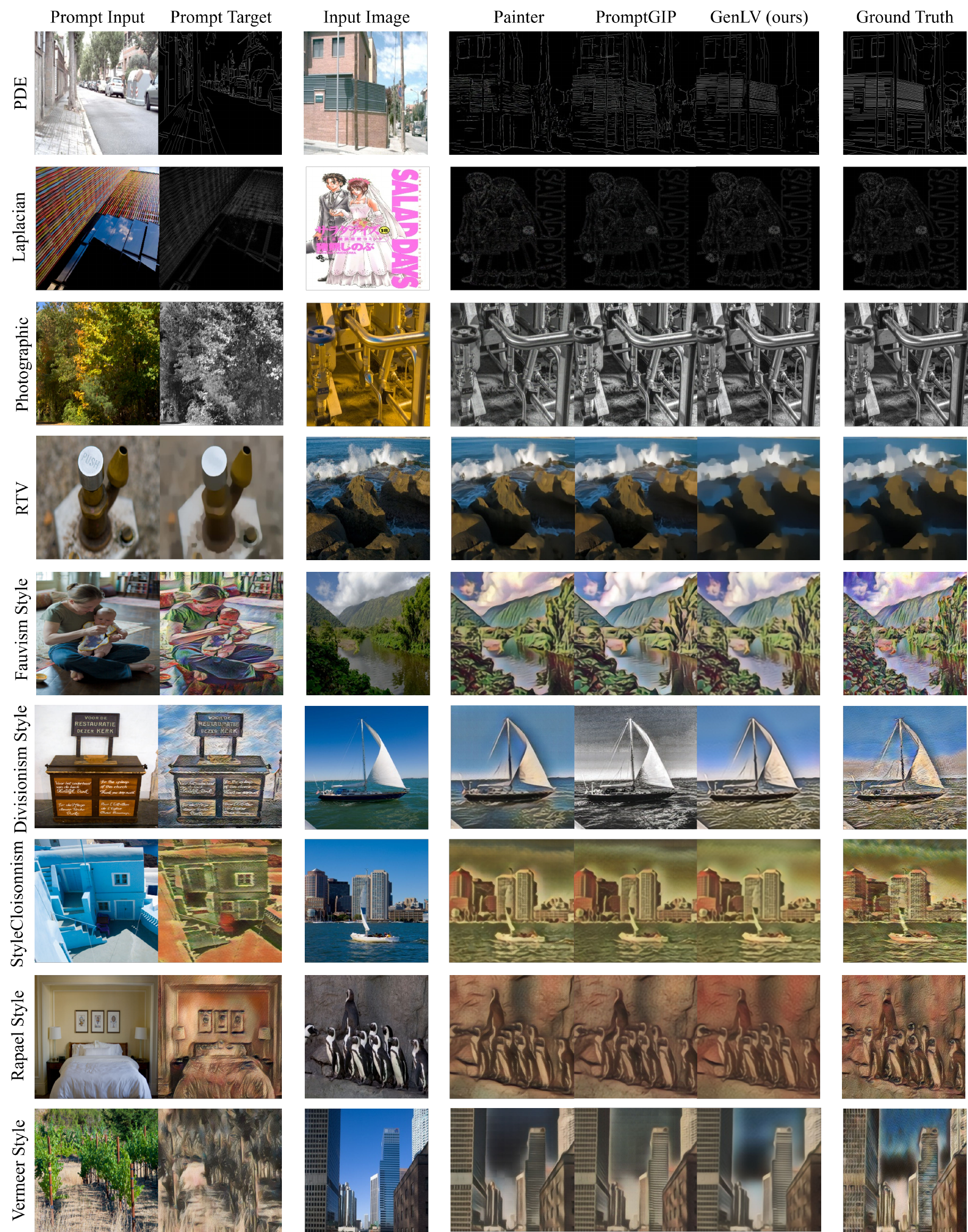}
    \caption{More visual results of different models on various low-level vision tasks.}
    \label{fig:visual4}
\end{figure*}

In Figure 5 of the main paper, we show the visual results of 9 representative tasks across different methods. In Figure~\ref{fig:visual2}, Figure~\ref{fig:visual3} and Figure~\ref{fig:visual4}, we present more visual results on the remaining 21 tasks. 
We can see that GenLV produces the best visual results in various low-level vision tasks, with the sharpest textures and no blocking artifacts and color distortions.

\begin{table*}[t]
\centering
\caption{Quantitative results (PSNR) of different image restoration backbone networks.}
\label{tab:backbone}
\resizebox{\linewidth}{!}{
\begin{tabular}{c|ccccccccccc}
\toprule
\rowcolor{color3} & GN & PN & S\&P Noise & GB & JPEG & Ringing & R-L & Inpainting & SimpleRain & ComplexRain & Haze     \\ 
\midrule
RRDB        & 26.05 & 27.42 & 24.85 & 22.77 & 25.37 & 24.51 & 25.01 & 24.28 & 24.20 & 22.69 & 21.54 \\
ViT         & 24.67 & 25.39 & 23.71 & 22.17 & 24.76 & 23.89 & 24.09 & 23.11 & 23.21 & 23.04 & 24.91 \\
SwinIR      & 28.83 & 31.19 & 36.59 & 23.45 & 26.65 & 26.00 & 29.51 & 27.00 & 29.78 & 22.26 & 21.23 \\
Restormer   & 28.56 & 31.21 & 35.42 & 24.16 & 26.65 & 27.00 & 29.83 & 27.77 & 29.38 & 24.16 & 14.83 \\
X-Restormer & 28.70 & 31.36 & 35.33 & 24.13 & 26.68 & 26.88 & 30.01 & 27.68 & 29.65 & 24.39 & 16.73 \\
\bottomrule
\end{tabular}
}
\end{table*}

\begin{table*}[t]
\centering
\caption{Quantitative results of using different prompt interaction mechanisms.}
\label{tab:modulation}
\resizebox{\linewidth}{!}{
\begin{tabular}{c|ccccccccccc}
\toprule
\rowcolor{color3} & GN & PN & S\&P Noise & GB & JPEG & Ringing & R-L & Inpainting & SimpleRain & ComplexRain & Haze     \\ 
\midrule
Without Prompt Interaction & 24.30 & 25.85 & 26.54 & 20.63 & 19.26 & 16.88 & 17.87 & 22.57 & 19.56 & 21.55 & 14.22 \\
Feature Modulation - GFM & 27.76 & 30.04 & 32.42 & 22.52 & 25.68 & 24.55 & 25.65 & 26.12 & 25.66 & 24.56 & 28.55 \\
Feature Modulation - SFT & 28.03 & 30.58 & 33.20 & 22.82 & 26.04 & 24.72 & 26.46 & 26.42 & 26.48 & 24.46 & 28.17 \\
Prompt Cross-Attention (ours) & 28.28 & 30.80 & 33.47 & 23.14 & 26.06 & 25.50 & 27.51 & 26.66 & 27.68 & 25.13 & 28.65 \\
\bottomrule
\end{tabular}
}
\end{table*}

\begin{table*}[t]
\centering
\caption{Comprehensive comparison between PromptGIP and GenLV under three different settings. $\star$: trained only for restoration tasks. $\#$: trained on the PromptGIP setting (15 tasks).  $\dagger$: trained on the GenLV setting (30 tasks).}
\label{tab:comparison}
\resizebox{\linewidth}{!}{
\begin{tabular}{c|ccccccccccc}
\toprule
\rowcolor{color3} & GN & PN & S\&P Noise & GB & JPEG & Ringing & R-L & Inpainting & SimpleRain & ComplexRain & Haze     \\ 
\midrule
PromptGIP\textsuperscript{$\star$}   & 26.48 & 27.76 & 28.08 & 22.88 & 25.86 & 25.69 & 27.05 & 25.28 & 25.79 & 24.33 & 24.55 \\
GenLV\textsuperscript{$\star$}(ours) & 28.99 & 31.69 & 36.63 & 24.58 & 26.91 & 27.74 & 31.50 & 28.11 & 31.10 & 24.71 & 28.91 \\
\hline
PromptGIP\textsuperscript{$\#$}   & 26.22 & 27.29 & 27.49 & 22.77 & 25.38 & 25.45 & 26.79 & 25.02 & 25.46 & 24.08 & 24.32 \\
GenLV\textsuperscript{$\#$}(ours) & 28.92 & 31.58 & 36.32 & 24.33 & 26.55 & 27.55 & 31.11 & 27.86 & 30.35 & 24.47 & 28.73 \\
\hline
PromptGIP\textsuperscript{$\dagger$}   & 23.63 & 23.98 & 25.05 & 20.84 & 22.21 & 23.86 & 24.94 & 22.11 & 23.16 & 21.79 & 21.90 \\
GenLV\textsuperscript{$\dagger$}(ours) &
28.49 & 31.05 & 34.20 & 23.39 & 26.21 & 25.78 & 28.21 & 27.17 & 28.18 & 25.11 & 29.70 \\
\bottomrule
\end{tabular}
}
\end{table*}

\end{document}